# Prediction and Control with Temporal Segment Models

**Nikhil Mishra** [1] **Pieter Abbeel** [1 2] **Igor Mordatch** [2]

## Abstract

We introduce a method for learning the dynamics of complex nonlinear systems based on deep generative models over *temporal segments* of states and actions. Unlike dynamics models that operate over individual discrete timesteps, we learn the distribution over future state trajectories conditioned on past state, past action, and planned future action trajectories, as well as a latent prior over action trajectories. Our approach is based on convolutional autoregressive models and variational autoencoders. It makes stable and accurate predictions over long horizons for complex, stochastic systems, effectively expressing uncertainty and modeling the effects of collisions, sensory noise, and action delays. The learned dynamics model and action prior can be used for end-to-end, fully differentiable trajectory optimization and model-based policy optimization, which we use to evaluate the performance and sample-efficiency of our method.

## 1. Introduction

The problem of learning dynamics – where an agent learns a model of how its actions will affect its state and that of its environment – is a key open problem in robotics and reinforcement learning. An agent equipped with a dynamics model can leverage model-predictive control or model-based reinforcement learning (RL) to perform a wide variety of tasks, whose exact nature need not be known in advance, and without additional access to the environment.

In contrast with model-free RL, which seeks to directly learn a policy (mapping from states to actions) in order to accomplish a specific task, learning dynamics has the advantage that dynamics models can be learned without task-specific supervision. Since dynamics models are decoupled from any particular task, they can be reused across different tasks in the same environment. Additionally, learning differentiable dynamics models (such as those based on neural networks) enables the use of end-to-end backpropagation-based methods for policy and trajectory optimization that are much more efficient than model-free methods.

Typical approaches to dynamics learning build a one-step model of the dynamics, predicting the next state as a function of the current state and the current action. However, when chained successively for many timesteps into the future, the predictions from a one-step model tend to diverge from the true dynamics, either due to the accumulation of small errors or deviation from the regime represented by the data the model was trained on. Any learned dynamics model is only valid under the distribution of states and actions represented by its training data, and one-step models make no attempt to deal with the fact that they cannot make accurate predictions far outside this distribution.

When the true dynamics are stochastic, or the sensory measurements noisy or unreliable, these problems are only exacerbated. Moreover, the dynamics may be inherently difficult to learn: bifurcations such as collisions induce sharp changes in state that are hard to model with certainty when looking at a single timestep. There may also be hysteresis effects such as gear backlash in robots, or high-order dynamics in hydraulic robot actuators and human muscles that require looking at a history of past states.

We present a novel approach to learning dynamics based on a deep generative model over temporal segments: we wish to model the distribution over possible future state trajectories conditioned on planned future actions and a history of past states and actions. By considering an entire segment of future states, our approach can model both uncertainty and complex interactions (like collisions) holistically over a segment, even if it makes small errors at individual timesteps. We also model a prior over action segments using a similar generative model, which can be used to ensure that the action distribution explored during planning is the same as the one the model was trained on. We show that that our method makes better predictions over long horizons than one-step models, is robust to stochastic dynamics and measurements, and can be used in a variety of control settings while only considering actions from the regime where the model is valid.

[1]University of California, Berkeley [2]OpenAI. Correspondence to: Nikhil Mishra <nmishra@berkeley.edu>.

## 2. Related Work

A number of options are available for representation of learned dynamics models, including linear functions (Mordatch et al., 2016; Yip & Camarillo, 2014), Gaussian processes (Boedecker et al., 2014; Ko & Fox, 2009; Deisenroth & Rasmussen, 2011), predictive state representations (PSRs) (Littman et al., 2002; Rosencrantz et al., 2004), and deep neural networks (Punjani & Abbeel, 2015; Fragkiadaki et al., 2015; Agrawal et al., 2016). Linear functions are efficient to evaluate and solve controls for, but have limited expressive power. Gaussian processes (Williams & Rasmussen, 1996) provide uncertainty estimates, but scaling them to large datasets remains a challenge (Shen et al.; Lawrence et al., 2003). PSRs and variants make multi-step predictions, but suffer from the same scalability challenges as Gaussian processes. Our method combines the expressiveness and scalability of neural networks with the ability to provide sampling and uncertainty estimates, modeling entire segments to improve stability and robustness.

An alternative is to learn dynamics models in an online fashion, constantly adapting the model based on an incoming stream of observed states and actions (Fu et al., 2016; Mordatch et al., 2016; Yip & Camarillo, 2014; Lenz et al., 2015). However, such approaches are slow to adapt to rapidly-changing dynamics modes (such as those arising when making or breaking contact) and may be problematic when applied on robots performing rapid motions.

Several approaches exist to improve the stability of models that make sequential predictions. Abbeel & Ng (2004) and Venkatraman et al. (2015) consider alternative loss functions that improve robustness over long prediction horizons. Bengio et al. (2015) and Venkatraman et al. (2016) also use simple curricula for a similar effect. While they all consider multi-step prediction losses, they only do so in the context of training models that are intrinsically one-step.

Existing methods for video prediction (Finn & Levine, 2016; Oh et al., 2015) look at a history of previous states and actions to predict the next frame; we take this a step further by modeling a distribution over an entire segment of future states that is also conditioned on future actions. In this work, we focus on demonstrating the benefits of a probabilistic segment-based approach; these methods could easily be incorporated with ours to learn dynamics from images, but we leave this to future work.

Watter et al. (2015) and Johnson et al. (2016) use variational autoencoders to learn a low-dimensional latent-space representation of image observations. Finn et al. (2016) takes a similar approach, but without the variational aspect. These works utilized autoencoders as a means of dimensionality reduction (rather than for temporal coherence like we do) to enable the use of existing control algorithms based on locally-linear one-step dynamics models.

Temporally-extended actions were shown to be effective in reinforcement learning, such as the options framework (Sutton et al., 1999b) or sequencing of sub-plans (Vezhnevets et al., 2016). Considering entire trajectories as opposed to single timesteps can also lead to simpler control policies. For example, there are effective and simple manually-designed control laws (Raibert, 1986), (Pratt et al., 2006) that formulate optimal actions as a function of the entire future trajectory rather than a single future state.

## 3. Segment-Based Dynamics Model

Suppose we have a non-linear dynamical system with states $x_t$ and actions $u_t$. The conventional approach to learning dynamics is to learn a function $x_{t+1} = f(x_t, u_t)$ using an approximator such as a neural network (possibly recurrent).

We consider a more general formulation of the problem, which is depicted in Figure 1: given segments (of length $H$) of past states $X^- = \{x_{t-H}, \ldots, x_{t-1}\}$ and actions $U^- = \{u_{t-H}, \ldots, u_{t-1}\}$, we wish to predict the entire segment of future states $X^+ = \{x_t, \ldots, x_{t+H}\}$ that would result from taking actions $U^+ = \{u_t, \ldots, u_{t+H-1}\}$. Treating these four temporal segments as random variables, then we wish to learn the conditional distribution $P(X^+|X^-, U^-, U^+)$. We introduce dependency on past actions $U^-$ to support dynamics with delayed or filtered actions.

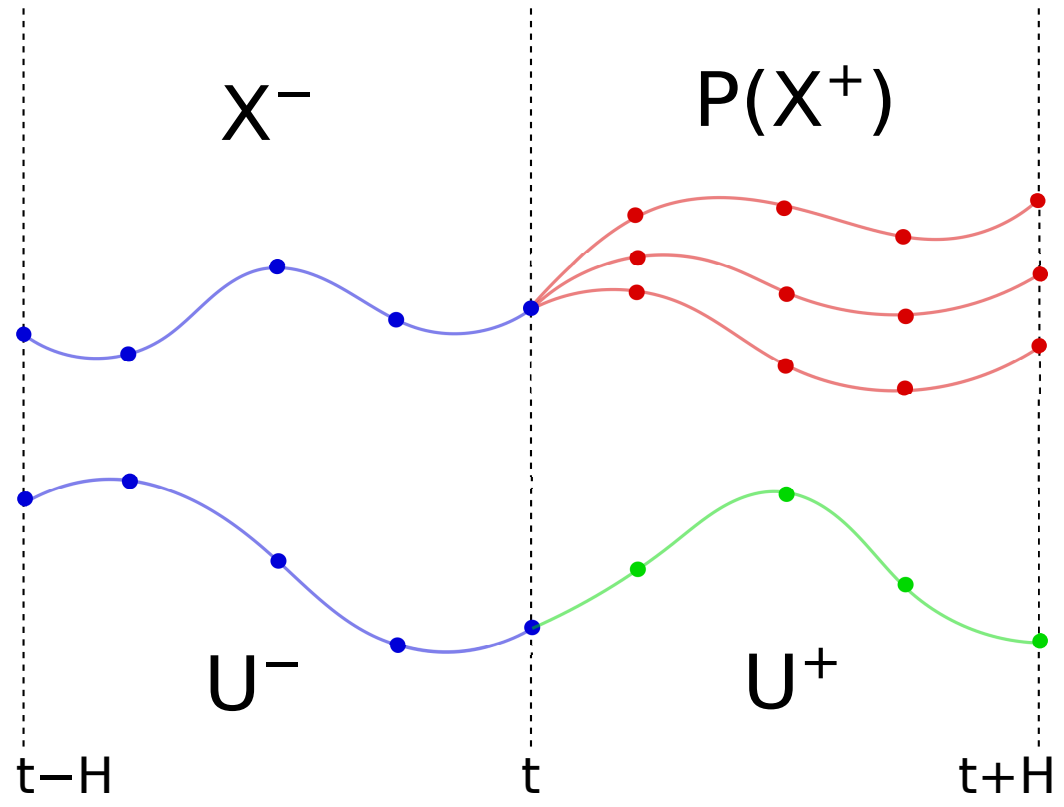

*Figure 1.* An overview of the probabilistic model we wish to learn. Given observed past states and actions $X^-, U^-$ (blue), and planned future actions $U^+$ (green), we wish to sample possible future state trajectories $X^+$ (red).

With this in mind, we propose the use of a deep conditional variational autoencoder (Kingma & Welling, 2014): our encoder will learn the distribution $Q^{(x)}(Z|X^+, X^-, U^-, U^+)$ over latent codes $Z$, and our decoder will learn to reconstruct $X^+$ from $X^-, U^-, U^+$ and a sample from $Z$, modeling the distribution $P^{(x)}(X^+|X^-, U^-, U^+, Z)$. Note that the random variable $Z$ is a vector that describes an entire segment of states, $X^+$. After training is complete, we can discard the encoder, and the decoder will allow us to predict the future state trajectory $\hat{X^+}$ using $X^-, U-, U^+$, as desired, sampling latent codes from an enforced prior $P(Z) = \mathcal{N}(0, I)$. Empirically, we observe that having the encoder model $Q^{(x)}(Z|X^+)$ instead of $Q^{(x)}(Z|X^+, X^-, U^-, U^+)$ gives equivalent performance, and so we take this approach in all of our experiments for simplicity.

### 3.1. Model Architecture and Training

In the previous section we discussed a conditional variational autoencoder whose generative path serves as a stochastic dynamics model. Here we will expand on some of the architectural details. A diagram of the entire training setup is shown in Figure 2. For more details of the architectures used in our experiments, see Appendix A.

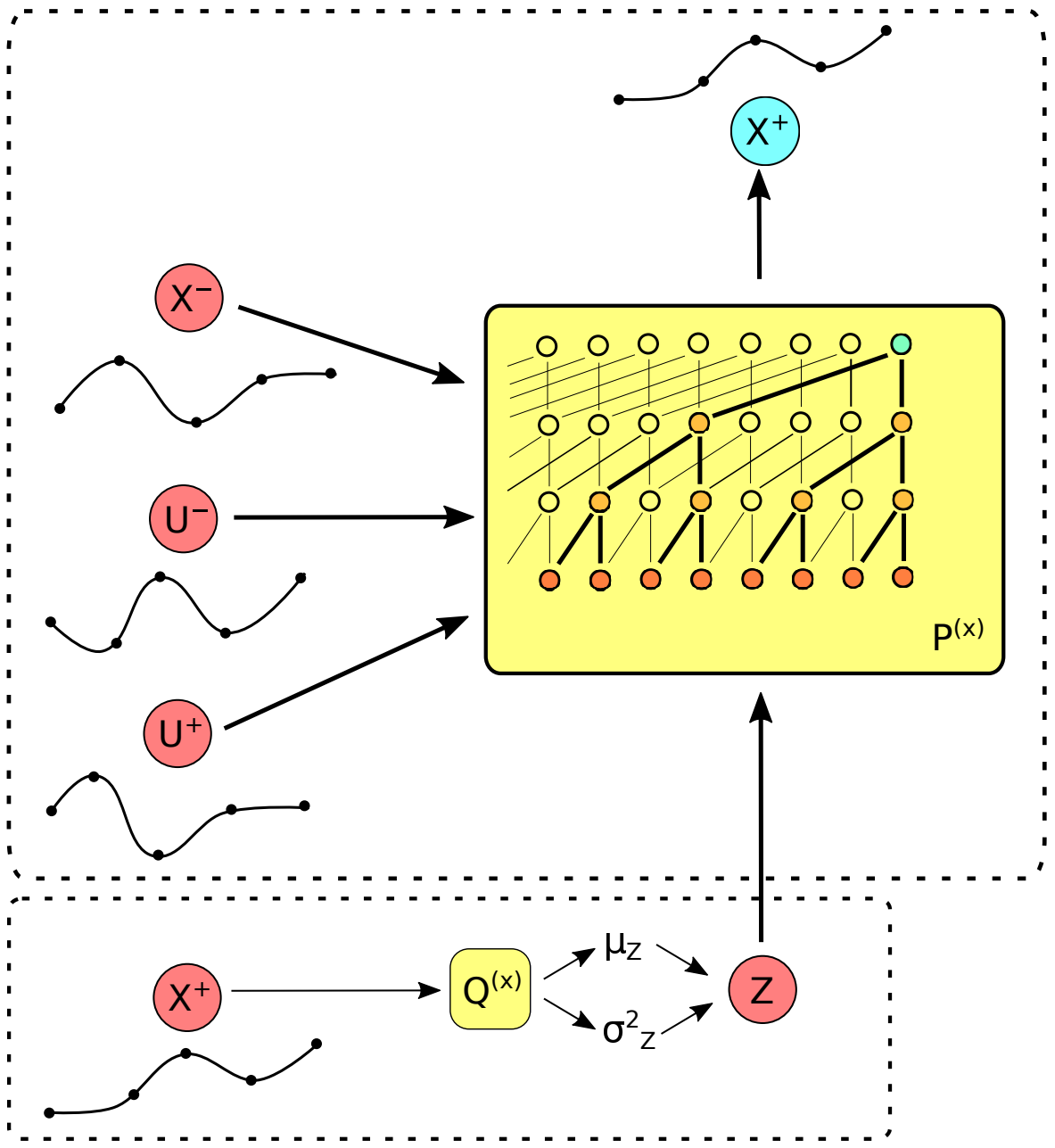

*Figure 2.* The dynamics model that we learn. The encoder $Q^{(x)}$ parametrizes a diagonal Gaussian distribution $Z \sim \mathcal{N}(\mu_Z, \sigma_Z^2)$ over latent codes describing a state trajectory $X^+$. The autoregressive decoder $P^{(x)}$ takes in segments of past states $X^-$, past actions $U^-$, and future actions $U^+$, along with a sample from $Z$, and uses dilated causal convolutions to reconstruct $X^+$.

The encoder network $Q^{(x)}$ explicitly parametrizes a Gaussian distribution over latent codes $Z$ with diagonal covariance. It consists of a stack of 1D-convolutional layers, whose output is flattened and projected into a single vector containing a mean $\mu_Z$ and variance $\sigma_Z^2$. We then sample $z \sim \mathcal{N}(\mu_Z, \sigma_Z^2)$ in a differentiable manner using the reparametrization trick (Kingma & Welling, 2014).

The decoder network $P^{(x)}$ seeks to model a distribution over a segment of states $P(X^+) = P(x_t, \ldots, x_{t+H})$. The causal nature of this segment (a particular timestep is only affected by the ones that occur before it) suggests that an autoregressive model with dilated convolutions is appropriate, similar to architectures previously used for modeling audio (van den Oord et al., 2016a) and image (van den Oord et al., 2016b) data. Like these works, we use layers with the following activation function:

$$\tanh(W_{f,k} * s + V_{f,k}^T z) \odot \sigma(W_{g,k} * s + V_{g,k}^T z) \qquad (1)$$

where $*$ denotes convolution, $\odot$ denotes elementwise multiplication, $\sigma(\cdot)$ is the sigmoid function, $s$ is the input to the layer, $z$ is a latent code sampled from the output of the decoder, and $W, V$ are network weights to be learned. We found that residual layers and skip connections between layers give slightly better performance but are not essential.

We train the model parameters end-to-end, minimizing the $l_2$-loss between $X^+$ and its reconstruction $\hat{X^+}$, along with the KL-divergence of the latent code $Z \sim \mathcal{N}(\mu_Z, \sigma_Z^2)$ from $\mathcal{N}(0, I)$ similarly to Kingma & Welling (2014).

## 4. Control with Segment-Based Models

Once we have learned a dynamics model, we want to utilize it in order to accomplish different tasks, each of which can be expressed as reward function $r(x_t, u_t)$. Trajectory optimization and policy optimization are two settings where a dynamics model would commonly be used, and provide meaningful ways with which to evaluate a dynamics model.

### 4.1. Trajectory Optimization

In trajectory optimization, we wish to find a sequence of actions that can be applied to accomplish a particular instance of a task. Specifically, given a reward function $r$, we want to maximize the sum of rewards along the trajectory that results from applying the actions $u_1, \ldots, u_T$, beginning from an initial state $x_0$. This can be summarized by the following optimization problem:

$$\max_{u_1,\ldots,u_T} \mathbf{E}\left[\sum_{t=1}^{T} r(x_t, u_t)\right] \quad \text{with } x_t \sim P^{(x)}(x_t|x_{0:t-1}, u_{1:t}) \qquad (2)$$

where $r(x_t, u_t)$ is the reward received at time $t$, and $X = \{x_1, \ldots, x_T\}$ is the sequence of states that would result from taking actions $U = \{u_1, \ldots, u_T\}$ from initial state $x_0$, under dynamics model $P^{(x)}$. The expectation is taken over state trajectories sampled from the model.

### 4.2. Latent Action Priors

If we attempt to solve the optimization problem as posed in (2), the solution will often attempt to apply action sequences outside the manifold where the dynamics model is valid: these actions come from a very different distribution than the action distribution of the training data. This can be problematic: the optimization may find actions that achieve high rewards under the model (by exploiting it in a regime where it is invalid) but that do not accomplish the goal when they are executed in the real environment.

To mitigate this problem, we propose the use of another conditional variational autoencoder, this one over segments of actions. In particular, given sequences of past actions $U^- = \{u_{t-H}, \dots, u_{t-1}\}$, and future actions $U^+ = \{u_t, \dots, u_{t+H}\}$, we wish to model the the conditional distribution $P(U^+|U^-)$. The encoder learns $Q^{(u)}(Z|U^+)$ and the decoder learns $P^{(u)}(U^+|Z, U^-)$. We condition on $U^-$ to support temporal coherence in the generated action sequence. Like the dynamics model introduced in Section 3.1, the encoder uses 1D-convolutional layers, and the decoder is autoregressive, with dilated causal convolutions. The latent space that this autoencoder learns describes a prior over actions that can be used when planning with a dynamics model; hence we refer to this autoencoder over action sequences as a *latent action prior*.

To incorporate a latent action prior, we divide an action sequence $U = \{u_1, \dots, u_T\}$ into segments $U_1, \dots U_K$ of length $H$ (where $K$ is determined such that $T = HK$, and $U_0 = 0$). Then we can generate action sequences that are similar to the ones in our training set by sampling different latent codes $z_1, \dots, z_K$ and using the decoder to sample from $P^{(u)}(U_k|U_{k-1}, z_k), \forall k = 1, \dots, K$. The optimization problem posed in (2) can then be expressed as:

$$\begin{aligned} &\max_{z_1,\dots,z_K} \mathbf{E}\left[\sum_{t=1}^{T} r(x_t, u_t)\right] \\ &\text{with } x_t \sim P^{(x)}(x_t|x_{0:t-1}, u_{1:t}) \\ &\qquad u_t \sim P^{(u)}(u_t|u_{1:t-1}, z_{1:K}) \end{aligned} \tag{3}$$

where the actions $u_1, \dots, u_T$ and states $x_1, \dots, x_T$ are generated by the latent action prior and dynamics model (see Figure 3 for an illustration). Since the dynamics model is differentiable, the above optimization problem can be solved end-to-end with backpropagation. While it is still nonconvex, we are optimizing over fewer variables, and the possible action sequences that are explored will be from the same distribution as the model's training data. Moreover, the gradients of the rewards with respect to the latent codes are likely to have stronger signal than those with respect to a single action. We used Adam (Kingma & Ba, 2015) with step size 0.01 to perform this optimization and found that it generally converged in around 100 iterations.

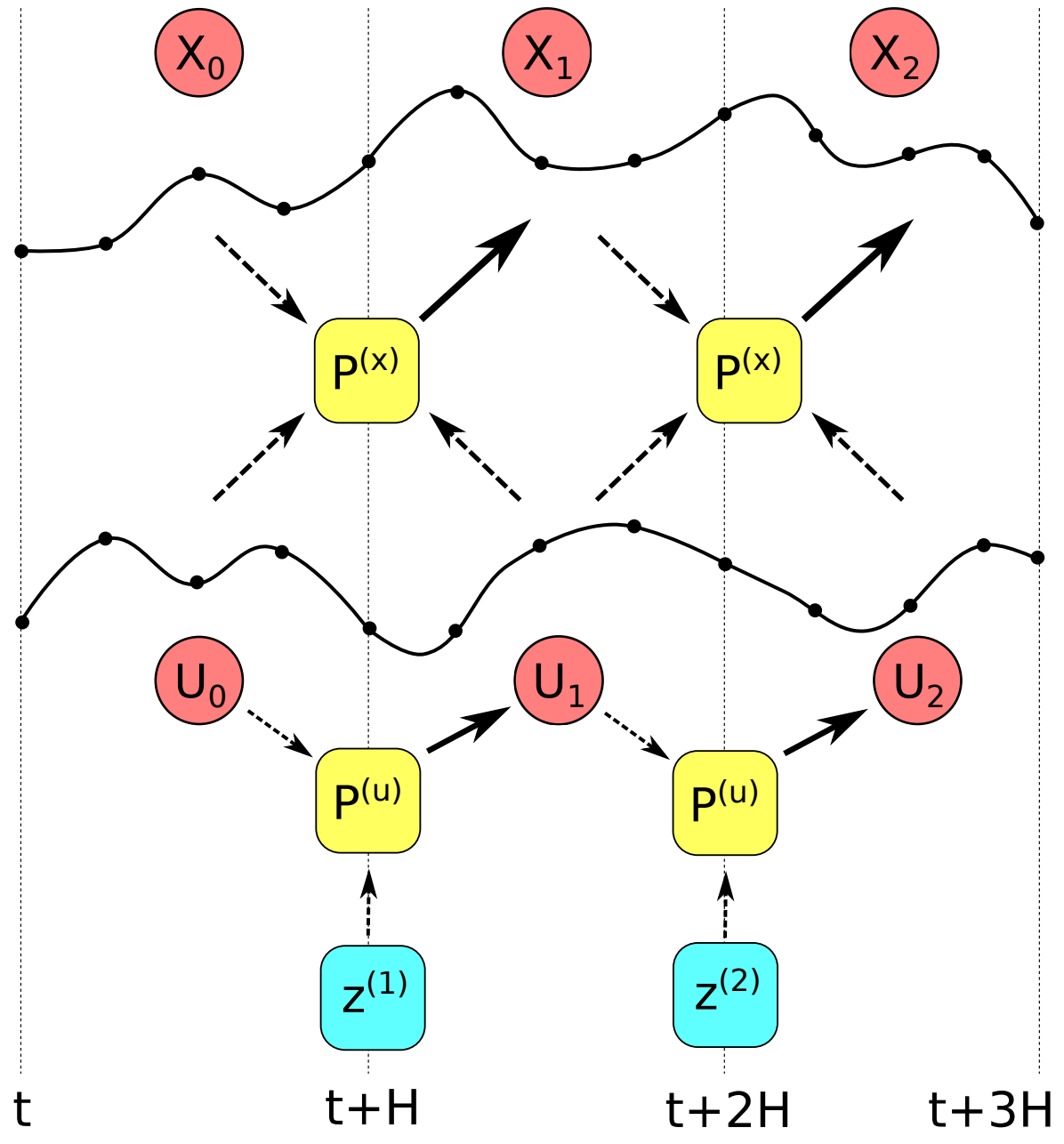

*Figure 3.* Trajectory optimization over latent codes (blue). The action sequences are generated using the latent codes and the latent action prior $P^{(u)}$, and the state trajectory using the generated actions and the dynamics model $P^{(x)}$.

## 4.3. Policy Optimization

Trajectory optimization enables an agent to accomplish a single instance of a task, but more often, we are interested in policy optimization, where the agent learns a policy that dictates optimal behavior in order to accomplish the task in the general case. In particular, a policy is a learned function (with parameters $\theta$) that defines a conditional distribution over actions given states, denoted $\pi_\theta(u|x)$. The value of a policy is defined as the expected sum of discounted rewards when acting under the policy, and can be expressed as:

$$\eta(\theta) = \mathbf{E}\left[\sum_{t=1}^{\infty} \gamma^t \cdot r(x_t, u_t)\right] \tag{4}$$

where the actions are sampled from $\pi_\theta(u_t|x_t)$, and $\gamma$ is a discount factor. The goal of policy optimization is to maximize the value of the policy with respect to its parameters.

The class of algorithms known as policy gradient methods (Sutton et al., 1999a; Peters & Schaal, 2006) attempt to solve this optimization problem without considering a dynamics model. They execute a policy $\pi_\theta$ to get samples $x_1, u_1, r_1, \dots, x_T, u_T, r_T$ from the environment, and then update $\theta$ to get an improved policy, relying on likelihood ratio methods (Williams, 1992) to estimate the gradient $\frac{\partial \eta}{\partial \theta}$ because they cannot directly compute the derivatives of the rewards $r(x_t, u_t)$ with respect to the actions $u_1, \dots, u_t$.

Model-based policy optimization can be more efficient than traditional policy-gradient methods, because the gradient $\frac{\partial \eta}{\partial \theta}$ can be computed directly by backpropagation through a differentiable model. However, its success hinges on the accuracy of the dynamics model, as the optimization can exploit flaws in the model in the same way as discussed in Section 4.2. Heess et al. (2015) use a model-based approach where a one-step dynamics model is learned jointly with a policy in an online manner. To evaluate the robustness of our models, we experiment with learning policies offline, where the dynamics model is learned through unsupervised exploration of the environment, and no environment interaction is allowed beyond this exploration.

Instead of a one-step policy of the form $\pi_\theta(u_t|x_t)$, we also explored using a segment-based policy $\pi_\theta(Z|X^-, U^-)$ that generates actions using latent action prior $P^{(u)}$ as follows:

$$X^-, U^- = \{x_{t-H}, \ldots, x_{t-1}\}, \{u_{t-H}, \ldots, u_{t-1}\}$$
$$\text{sample } Z \sim \pi_\theta(Z|X^-, U^-)$$
$$\{u_t, \ldots, u_{t+H}\} = U^+ \sim P^{(u)}(U^+|Z, U^-)$$

and then acts according to action $u_t$. The resulting policy will learn to accomplish the task while only considering actions for which the dynamics model is valid. In terms of the options framework (Sutton et al., 1999b), we can think of this policy as considering a continuous spectrum of options, all of which are consistent with both past observed states and actions, and the data distribution under which the dynamics model makes good predictions.

## 5. Experiments

Our experiments investigate the following questions:

(i) How well do segment-based models predict dynamics?

(ii) How does prediction accuracy transfer to control applications? How does this scale with the difficulty of the task and stochasticity in the dynamics?

(iii) How is this affected by the use of latent action priors?

(iv) Is there any meaning or structure encoded by the latent space learned by the dynamics model?

### 5.1. Environments

In order for a dynamics model to be versatile enough for use in control settings, the training data needs to contain a variety of actions that explore a diverse subset of the state space. Efficient exploration strategies are an open problem in reinforcement learning and are not the focus of this work. With this in mind, we base our experiments on a simulated 2-DOF arm moving in a plane (as implemented in the Reacher environment in OpenAI Gym), because performing random actions in this environment results in sufficient exploration. We consider the following environments throughout our experiments (illustrated in Figure 4):

(i) The basic, unmodified Reacher environment.

(ii) A version containing an obstacle that the arm can collide with: the obstacle cannot move, but its position is randomly chosen at the start of each episode.

(iii) A version in which the arm can push a damped cylindrical object around the arena.

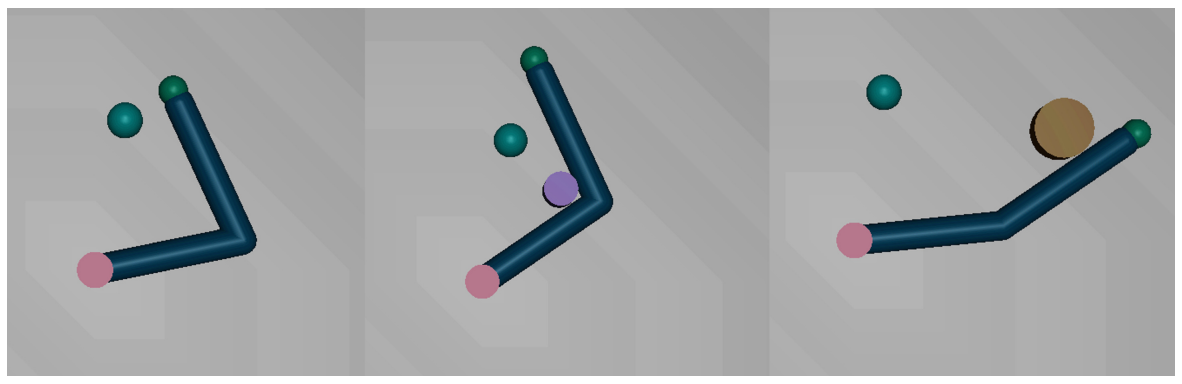

*Figure 4.* The environments we used in our experiments. From left to right: (i) the unmodified Reacher environment, (ii) a version with an obstacle, (iii) a version with an object to push. The blue marker is used to visualize the goal during experiments involving control but has no effect on the dynamics.

To learn a dynamics model, the training data consists of a collection of trajectories $x_1, u_1, \ldots, x_T, u_T$ from the environment we wish to model. We used 500 trajectories in the basic environment, and 5000 in the other two. For all environments, the state representation consists of the joint angles, the joint velocities, and the end-effector position; when obstacles are present, their positions and velocities are also included. The actions are always the torques for the arm's joints. In all experiments, the training set is comprised of trajectories of length $T = 100$ of the arm executing smooth random torques, and we used segments of length $H = 10$ and 8-dimensional latent spaces.

While the segment length and dimensionality of the latent space could be varied, we found that these values were reasonable choices for these environments. As the segment length approaches 1, the model degenerates into a one-step model, and for longer segments, its performance plateaus because the states towards the end of the segment become independent of those at the beginning. Likewise, we observed that this latent-space dimensionality was a good trade-off between expressiveness and information density.

### 5.2. Baselines

We compare our method against the following baselines:

(i) A one-step model: a learned function $x_{t+1} = f(x_t, u_t)$, where $f$ is a fully-connected neural network. It is trained using a one-step-prediction $l_2$-loss on tuples $(x_t, u_t, x_{t+1})$.

(ii) A one-step model that is rolled out several timesteps at training time. The model is still a learned function $x_{t+1} = f(x_t, u_t)$, but it is trained with a multi-step prediction loss, over a horizon of length $2H$. While this does not increase the model's expressive power, we expect it to be more robust to the accumulation of small errors (e.g., Venkatraman et al. (2015); Abbeel & Ng (2004)).

(iii) An LSTM model, which can store information about the past in a hidden state $h_t$: $x_{t+1}, h_{t+1} = f(x_t, u_t, h_t)$, and is trained with the same multi-step prediction loss (also over a horizon of $2H$). We expect that the LSTM can learn fairly complex dynamics, but the hidden state dependencies can make trajectory and policy optimization more difficult.

### 5.3. Results

#### 5.3.1. Dynamics Prediction

After learning a dynamics model, we evaluate it on a test set of held-out trajectories by computing the average log-likelihood of the test data under the model.

For our method, we do this by obtaining samples from the model, fitting a Gaussian to the samples, and determining the log-likelihood of the true trajectory under the fitted Gaussian. Since the baseline methods do not express uncertainty, but are trained using $l_2$-loss, we interpret their predictions as the mean of a Gaussian distribution whose variance is constant across all state dimensions and timesteps (since minimizing $l_2$-loss is equivalent to maximizing this log-likelihood). We then fit the value of the variance constant to maximize the log-likelihood on the test set.

Figure 5 compares our method to the baselines in each environment. The values reported are log-likelihoods per timestep, averaged over a test set of 1000 trajectories. Our model and the LSTM are competitive in the basic environment (and both substantially better than the one-step models), but the LSTM's performance degrades on more challenging environments with collisions.

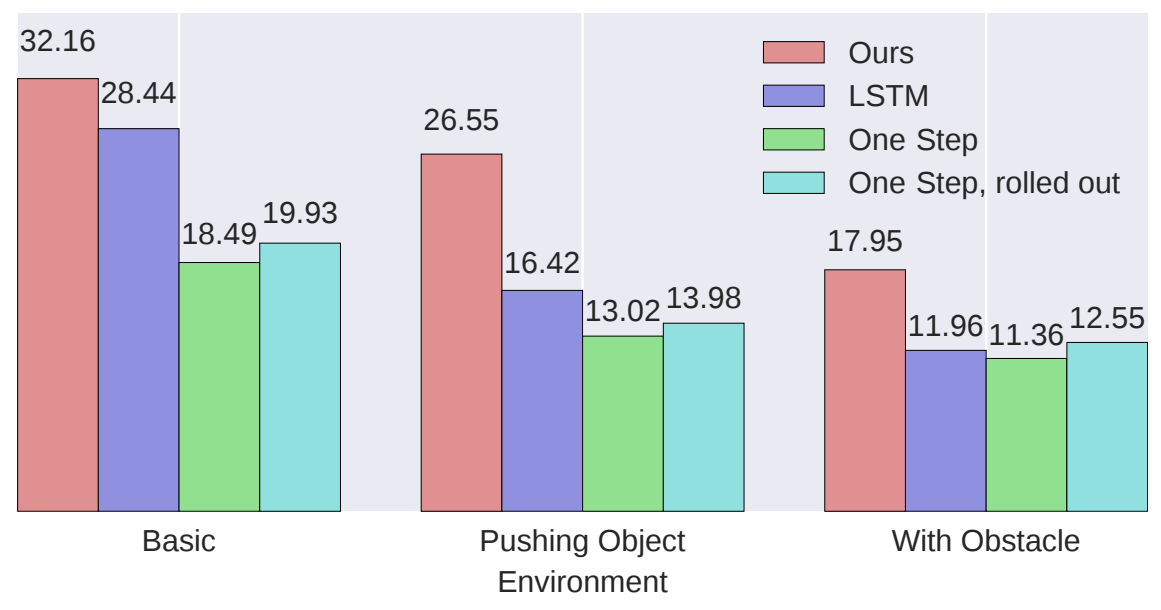

*Figure 5.* Prediction quality of our method compared to several baselines in a range of environments. The reported values are the average log-likelihood per timestep on a test set (higher is better). Our method significantly outperforms the baseline methods, even in environments with complex dynamics such as collisions.

#### 5.3.2. Control[1]

Next, we compare our method to the baselines on trajectory and policy optimization. Of interest is both the actual reward achieved in the environment, and the difference between the true reward and the expected reward under the model. If a control algorithm exploits the model to predict unrealistic behavior, then the latter will be large.

We consider two tasks:
(i) Reaching Task: the arm must move its end effector to a desired position. The reward function is the negative distance between the end effector and the target position, minus a quadratic penalty on applying large torques.

(ii) Pushing Task: the arm must push a cylindrical object to the desired position. Like in the reaching task, the reward function is the negative distance between the object and the target, again minus a penalty on large torques.

The trajectory-optimization results are summarized in Figure 6. For each task and dynamics model, we sampled 100 target positions uniformly at random, solved the optimization problem as described in (2) or (3), and then executed the action sequences in the environment in open loop.

Under each model, the optimization finds actions that achieve similar model-predicted rewards, but the baselines suffer from large discrepancies between model prediction and the true dynamics. Qualitatively, we notice that, on the pushing task, the optimization exploits the LSTM and one-step models to predict unrealistic state trajectories, such as the object moving without being touched or the arm passing through the object instead of colliding with it. Our model consistently performs better, and, with a latent action prior, the true execution closely matches the model's prediction. When it makes inaccurate predictions, it respects physical invariants, such as objects staying still unless they are touched, or not penetrating each other when they collide.

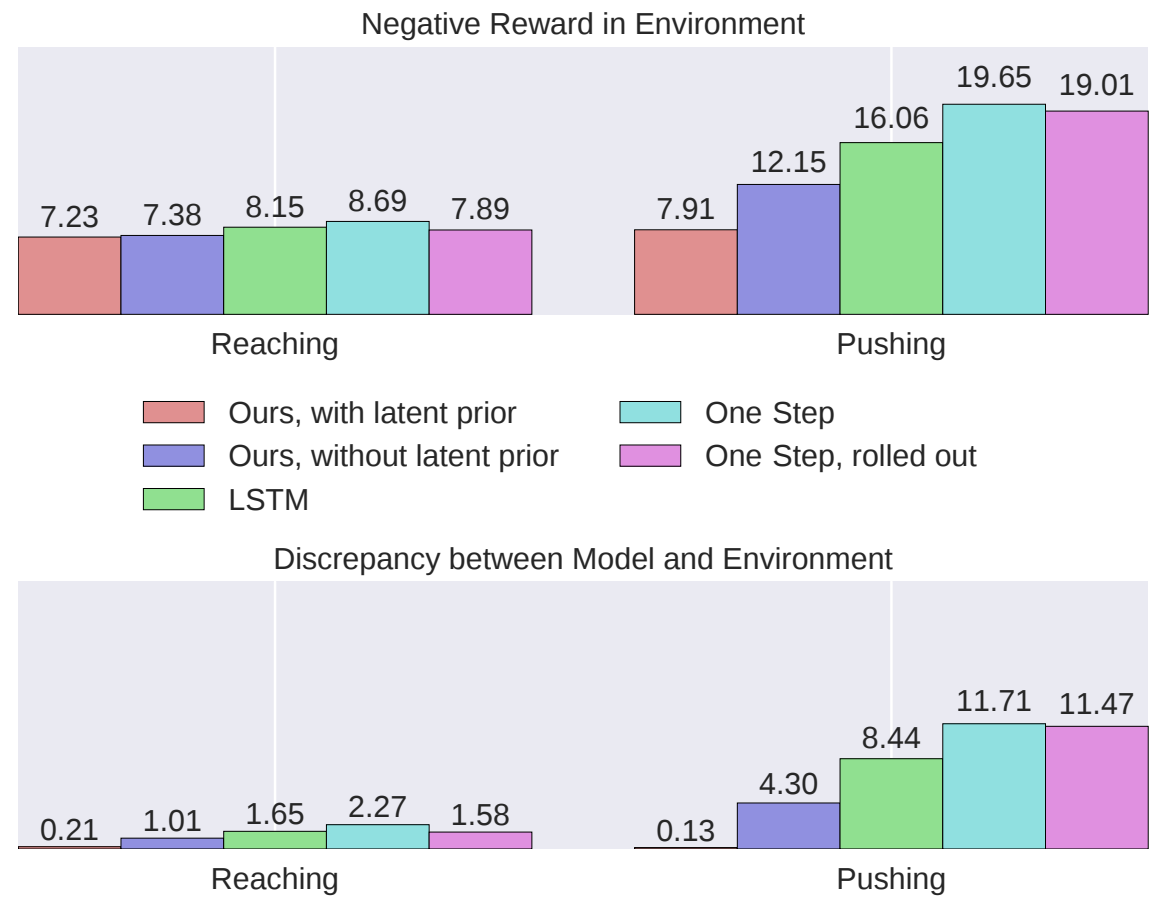

*Figure 6.* Trajectory optimization on the reaching and pushing tasks. The top plot reports the negative reward from open-loop execution of the returned action sequences (lower is better, averaged over 100 trials), and the bottom shows the difference between true reward and model-predicted reward. Our model, with a latent action prior, achieves both the best in-environment performance and the smallest discrepancy between environment and model.

Figure 7 depicts the results from policy-optimization (Section 4.3) in the form of learning curves for each task and dynamics model. See Appendix A for model architectures and hyperparameters. For comparison, we also plot the performance of a traditional policy gradient method. Although this method and ours eventually achieve similar performance, ours does so much more efficiently, learning the policy offline with fewer samples from the model than the traditional method needed from the environment.

[1] Videos of our experimental results can be seen here: https://sites.google.com/site/temporalsegmentmodels/.

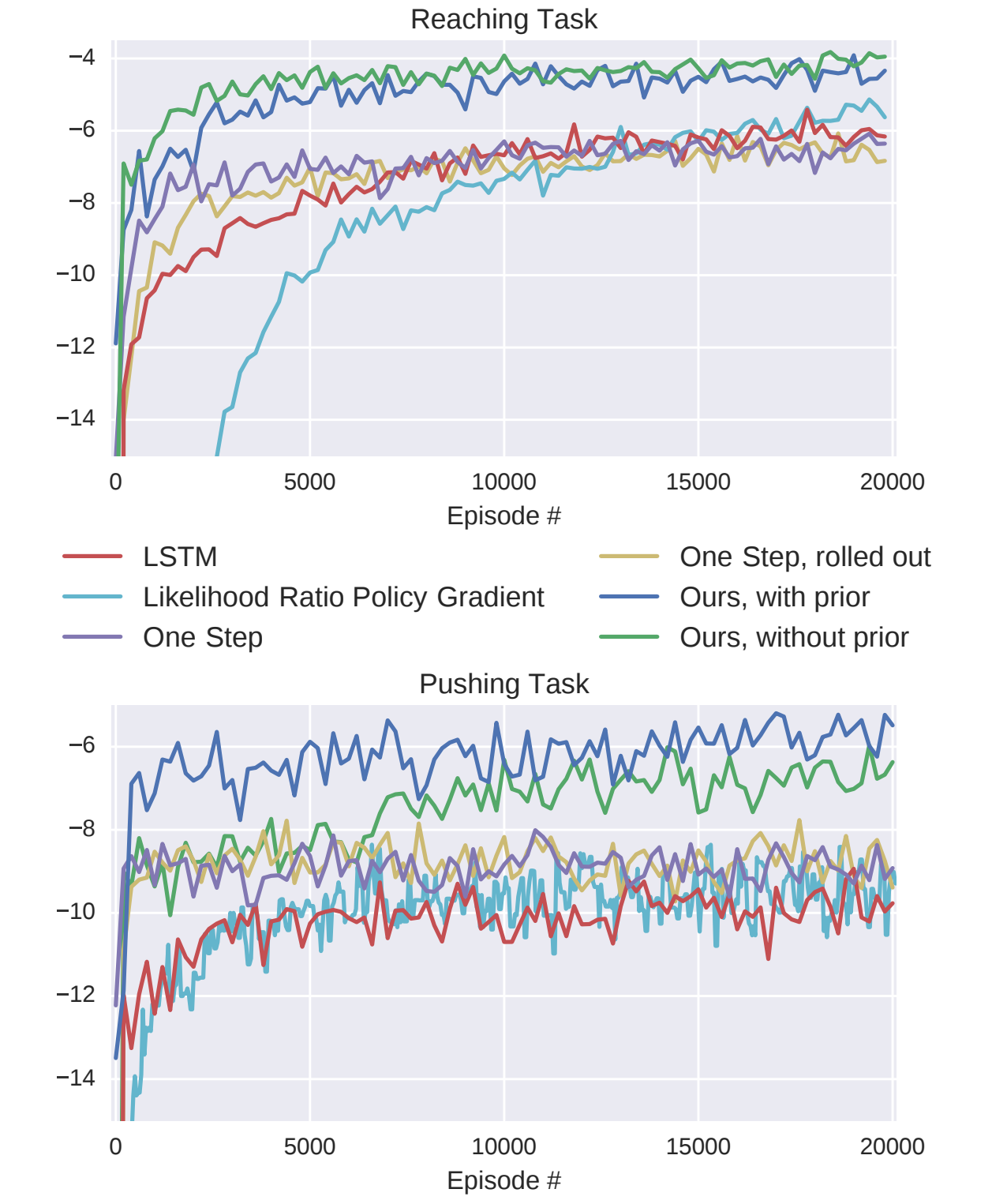

*Figure 7.* Learning curves of policy optimization on the reaching and pushing tasks (top and bottom, respectively). The quantities plotted are the true performance in the environment (100-episode average reward). Not only does our dynamics model, with an action prior, consistently perform the best, it is considerably faster than a model-free policy gradient method.

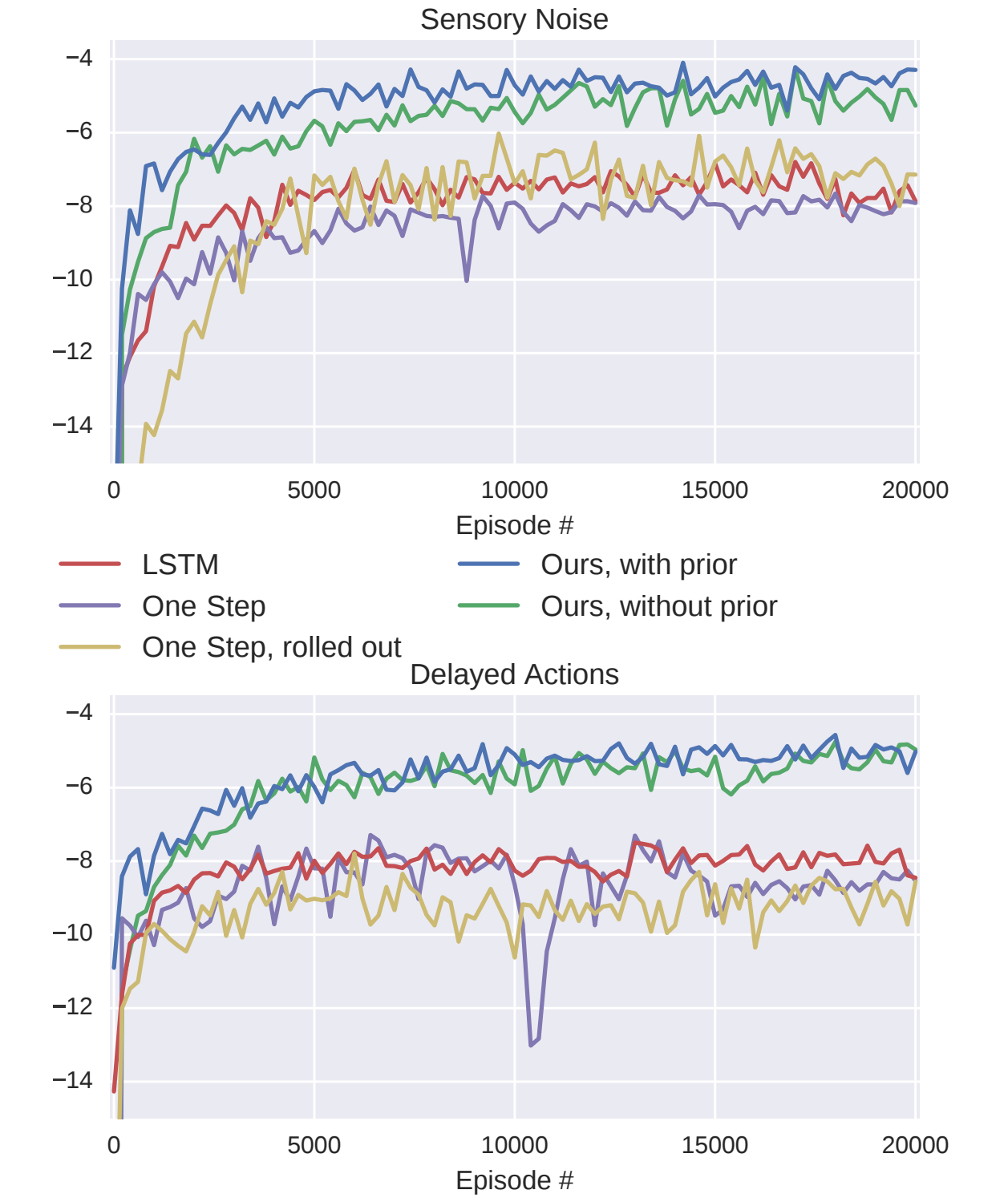

*Figure 8.* Our dynamics model, both with and without a latent action prior, can gracefully deal with noisy state observations and delayed actions, as depicted by these learning curves from the reaching task. Although the average rewards are slightly lower than in the absence of noise or delays, policies trained with the baseline models generally fail to perform the task.

#### 5.3.3. Sensory Noise and Delayed Actions

To explore the effects of stochastic dynamics and delayed actions, we consider two more modifications of the Reacher environment, one in which there is considerable Gaussian noise in the state observations ($\sigma = 0.25$ on data in the range $[-1, +1]$), and one in which actions are delayed: they do not take effect for $\tau = 5$ timesteps after they are applied. These challenges commonly arise in real-world robotics applications (Atkeson et al., 2016), and so it is important to be able to learn a useful dynamics model in either setting. For both the noisy-state and delayed-action environments, we learn a dynamics model with each method, and then use it to learn a policy for the reaching task. Figure 8 displays the resulting learning curves. Our dynamics model performs much better than the baselines, both with and without an action prior. Notably, using the LSTM model results in a substantially worse policy than ours even though its prediction accuracy is only slightly lower. Because our model operates over segments, it implicitly learns to filter noisy observations. This removes the need to explicitly apply and tune a filtering process, as is traditionally done.

#### 5.3.4. Analysis of Latent Space

Variational autoencoders are known for learning lossy latent codes that preserve high-level semantics of the data, leaving the decoder to determine the low-level details. As a result, we are curious to see whether our dynamics model learns a latent space that possesses similar properties.

Applied to dynamics data, one might expect a latent code to provide an overall description of what happens in the state trajectory $X^+$ it encodes. Alternatively, per the argument made by Chen et al. (2016), it is also conceivable that the decoder would ignore the latent code entirely, because the segments $X^-, U^-, U^+$ provide better information than $Z$ about $X^+$. However, we observe that our model does learn a meaningful latent space: one that encodes uncertainty about the future. A particular latent code corresponds to a particular future within the space of possible ones consistent with the given $X^-, U^-, U^+$.

When the dynamics are simple and deterministic (such as in the original Reacher environment), the model does express certainty by ignoring the latent code. With stochas-ticity (such as in the previous section), it provides a spread of reasonable state trajectories. Interestingly, when the dynamics are deterministic but complex, the model also uses the latent codes to express uncertainty. This can occur regarding the orientations and velocities of objects immediately following a collision, as illustrated in Figure 9.

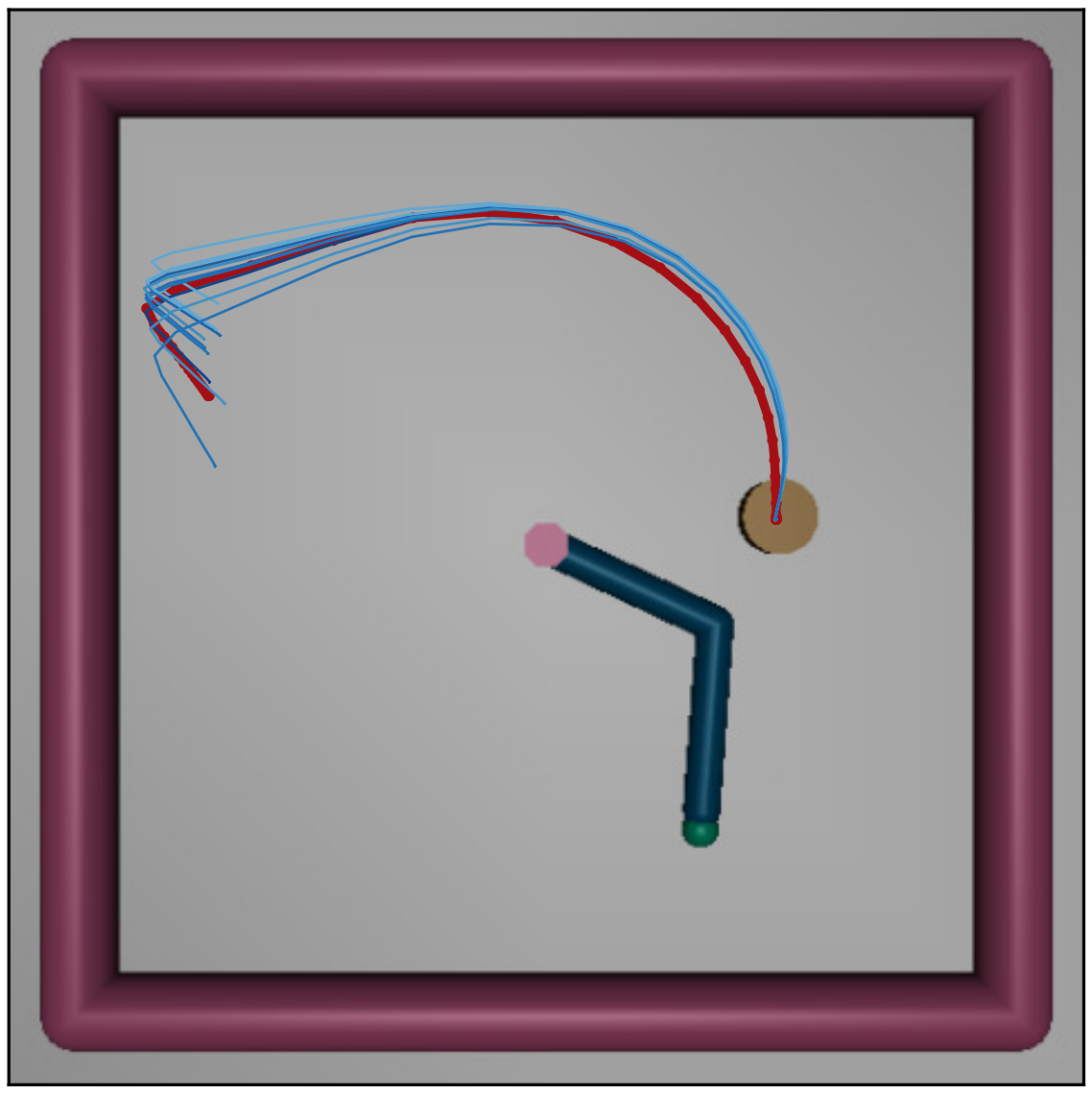

*Figure 9.* An episode from the pushing environment. The arm is about to swing counterclockwise and push the brown object; the red path indicates the observed motion of the object. The blue paths are samples from our model, given the same action sequence and initial state. It correctly predicts collisions between arm and object, and between object and wall, but expresses some uncertainty in the deflection angles and how far the object travels after bouncing off the wall.

### 5.3.5. Effect of Latent Action Prior

Our earlier experiments demonstrated the benefits of a latent action prior: by only considering actions for which the dynamics model is valid, the discrepancy between the model and the true dynamics is minimized, resulting in higher rewards achieved in the actual environment.

In this section, we qualitatively examine how the actions returned by control algorithms differ as a consequence of the latent action prior. An example is illustrated in Figure 10. In the training data, the actions that the agent takes are smooth, random torques, and we observe that when we use an action prior, solutions from trajectory optimization look similar. We contrast this with the solutions from optimizing directly over actions, which are sharp and discontinuous, unlike anything the dynamics model has seen before. This lets us infer that the baselines perform poorly on the pushing task (as shown in Figure 6) because of large discrepancies between the model prediction and the true execution.

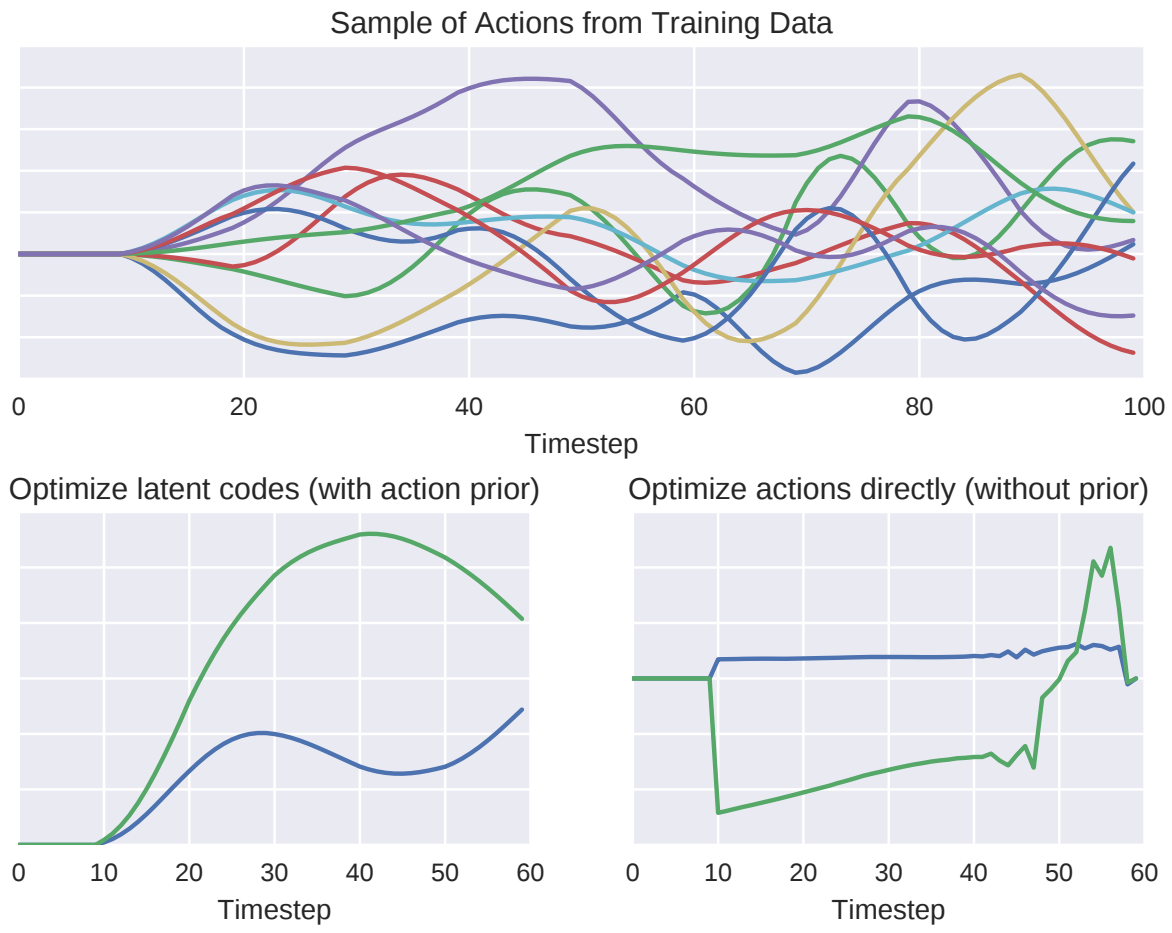

*Figure 10.* The qualitative effects of using a latent action prior, as seen during trajectory optimization. The top plot shows an example of action sequences from the training data. When we optimize over latent codes, the actions look similar (bottom left), but when we directly optimize over actions, the resulting sequence looks unlike anything the model has seen before.

## 6. Conclusion and Future Work

We presented a novel approach to dynamics learning based on temporal segments, using a variational autoencoder to learn the distribution over future state trajectories conditioned on past states, past actions, and planned future actions. We also introduced the latent action prior, a variational autoencoder that models a prior over action segments, and showed how it can be used to perform control using actions from the same distribution as a dynamics model's training data. Finally, through experiments involving trajectory optimization and model-based policy optimization, we showed that the resulting method can model complex phenomena such as collisions, is robust to sensory noise and action delays, and learns a meaningful latent space that expresses uncertainty about the future.

The most prominent direction for future work that we plan to explore, is the data collection procedure. In our experiments, correlated random actions resulted in sufficient exploration for the tasks we considered and allowed us to demonstrate the benefits of a segment-based approach. However, incorporating a more sophisticated exploration strategy to gather data (in an iterative procedure, potentially using the model's predictions to guide exploration) would allow us to tackle a more diverse set of environments, both simulated and real-world. The action prior and segment-based policy could be used as a starting point for hierarchical reinforcement-learning algorithms. Leveraging existing work on few-shot learning could help finetune a dynamics model during the policy learning process. Such approaches could yield significant advances in reinforcement learning, improving both sample efficiency and knowledge transfer between related tasks.

## Acknowledgements

Work done at Berkeley was supported in part by an ONR PECASE award.

# A. Appendix

Here we give a more precise description of the architectures of the models we introduced in the paper. Both the dynamics model and the latent action prior were trained using Adam with the default parameters.

## A.1. Dynamics Model

The following figure depicts the detailed encoder and decoder architectures for our dynamics models. The encoder uses 1D-convolutions (across the temporal dimension) and the ReLU activation function. The decoder is autoregressive, using dilated causal 1D-convolutions and the gated activation function described in Section 3.1.

The layer sizes indicated below correspond to the model we trained for the basic Reacher environment. For the obstacle and pushing environments, we used the same encoder architecture. The decoder for those environments had 64 channels in all layers, and had an additional $1 \times 1$ convolution with 128 channels before the final layer.

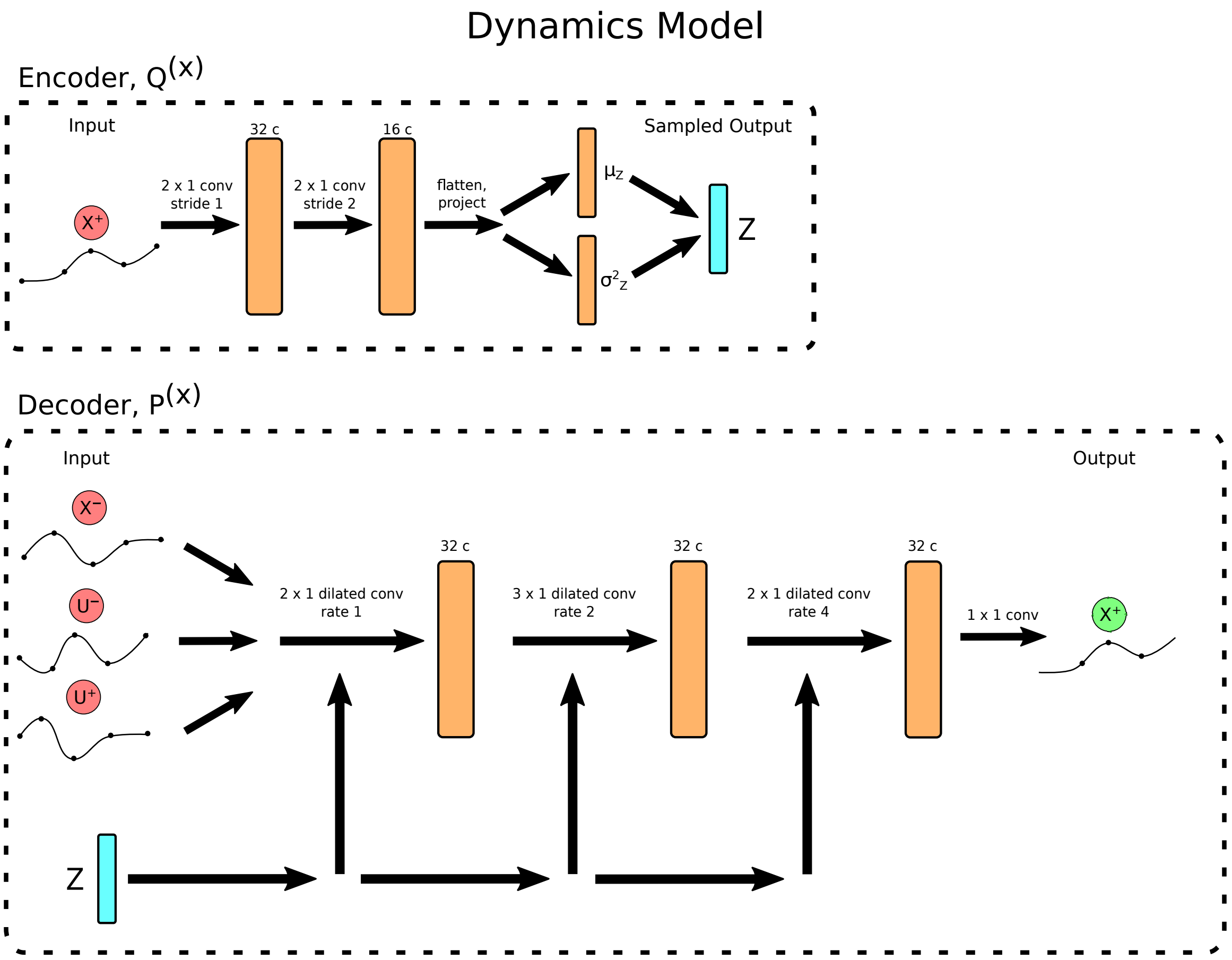

### A.2. Latent Action Prior

The architecture for the latent action prior is quite similar to that of our dynamics models as depicted on the previous page. We used the same architecture for all experiments.

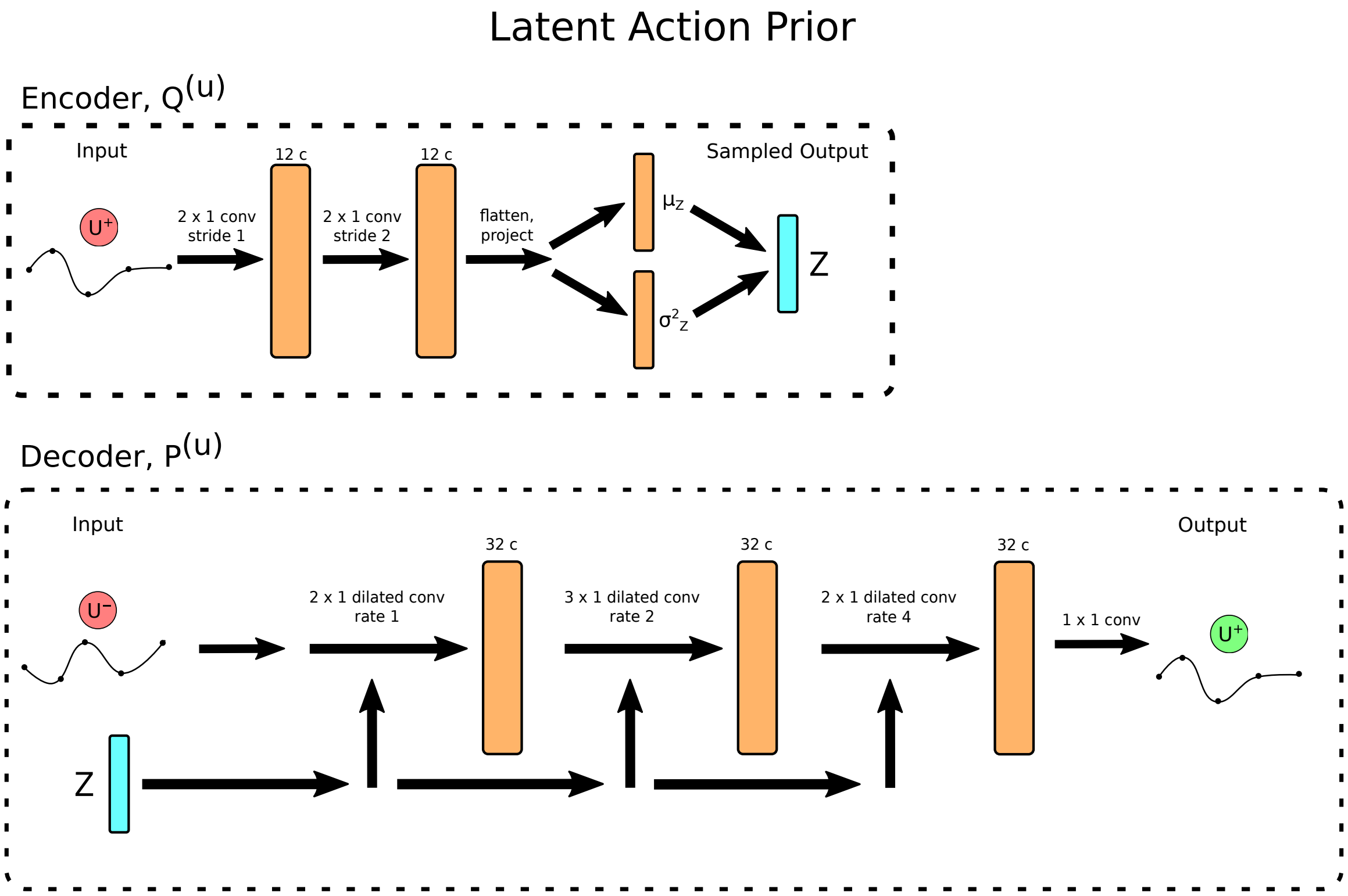

### A.3. Policy Optimization

The one-step policies had two hidden layers of size 64, and used the ReLU activation function.

The following figure illustrates the architecture of the segment-based policy, as introduced in Section 4.3. It uses the same gated activation function as the decoders of the other two models. The latent code that the policy produces is used by the action prior to generate a sequence of actions, of which the first one is executed.

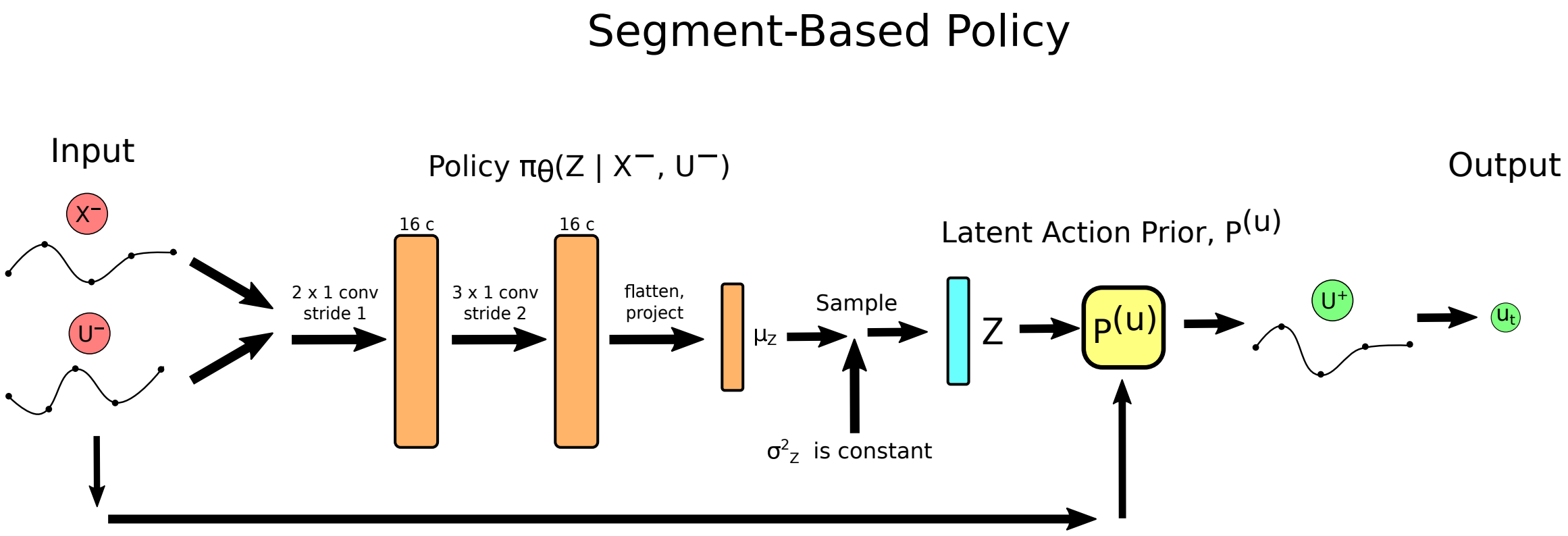

For all policy optimization experiments, we used a discount factor $\gamma = 0.99$, and computed the gradient of the policy's value with respect to its parameters (as discussed in Section 4.3) using backpropagation through time over 50 timesteps. We interpreted each policy's output as the mean of a diagonal Gaussian distribution, and sampled actions using a constant standard deviation of 0.01. Policy updates were computed using Adam over 20-episode batches, where the step size was initialized to $10^{-3}$ and decayed by a factor of 0.9 per 100 iterations until it reached $10^{-4}$ (all other parameters were left at the defaults).